\documentclass{article}
\usepackage{graphicx}

 \usepackage[preprint]{neurips_2025}
\workshoptitle{Generative AI in Finance}

\usepackage[utf8]{inputenc} 
\usepackage[T1]{fontenc}    
\usepackage{hyperref}       
\usepackage{url}            
\usepackage{booktabs}       
\usepackage{amsfonts}       
\usepackage{nicefrac}       
\usepackage{microtype}      
\usepackage{xcolor}         
\usepackage{amsmath}
\usepackage{wrapfig}

\title{Data-Efficient Realized Volatility Forecasting with Vision Transformers}

\author{%
  Emi Soroka \\
  Department of Electrical Engineering\\
  Stanford University\\
  Stanford, CA 94306 \\
  \texttt{esoroka@stanford.edu} \\
  \And
  Artem Arzyn \\
  Department of Electrical Engineering \\
  Stanford University\\
  Stanford, CA 94306 \\
  \texttt{arzyn@stanford.edu} \\
}

\begin{document}

\maketitle

\begin{abstract}
    Recent work in financial machine learning has shown the virtue of complexity: 
    the phenomenon by which deep learning methods capable of learning highly nonlinear 
    relationships outperform simpler approaches in financial forecasting. 
    While transformer architectures like Informer have shown promise for financial time series forecasting,
    the application of transformer models for options data remains largely unexplored. We conduct
    preliminary studies towards the development of a transformer model for options data by training the Vision Transformer (ViT) architecture, typically
    used in modern image recognition and classification systems, to predict the realized 
    volatility of an asset over the next 30 days from its implied volatility surface (augmented with date information) for a single day. We show that the ViT can learn seasonal patterns and nonlinear features from the IV surface, suggesting a promising direction for model development.
\end{abstract}

\section{Background}
The implied volatility surface (IV surface) of an optionable asset encodes information about market dynamics and sentiment, the future realized volatility of the asset, and the probability distribution of its return \cite{bali2022factor}. Traders construct features from the IV surface to infer this information using options pricing theory or empirical observations. Recent work in financial machine learning has also discovered the virtue of complexity:  \cite{empiricalassetpricing, complexityinfactormodels} the existence of highly nonlinear features in financial data which can be extracted using neural networks, contradicting prior assumptions that financial returns can be explained by a small number of predictive factors. However, machine learning methods are difficult to apply to financial data because the data itself is noisy and limited in scale. For example, our entire preprocessed dataset totals 6.1 GB while text corpora used to train frontier LLMs contain multiple terabytes of data \cite{liu2024datasetslargelanguagemodels}.

\section{Prior Work}
The use of neural networks to identify nonlinear patterns in financial data is investigated extensively in \cite{empiricalassetpricing}. Other examples include overparametrized factor models with more factors than assets under observation \cite{complexityinfactormodels}, Transformer-based time series forecasting \cite{zhou2021informer}, and structured approaches to machine learning in finance \cite{dixon2019fourhorsemen}. Neural networks have also been applied to generate smooth, arbitrage-free IV surfaces from raw option prices \cite{ackerer2020deepsmoothingimpliedvolatility, wiedemann2025operatordeepsmoothingimplied}. However, fewer researchers have investigated deep learning for predictions from IV surfaces. Previous approaches include the use of hand-constructed features \cite{Neuhierl2022Option} or convolutional neural networks (CNNs) \cite{deeplearningfromivsurfaces}, with the latter using the IV surface on the last trading day of the month to predict the monthly return of the next month. We train Vision Transformer (ViT) models on IV surfaces, treating them as small, single-channel images. ViTs on images require less computational cost than CNNs and provide more stable training performance \cite{dosovitskiy2021imageworth16x16words}; thus we hypothesize they will be more robust to noisy data and outliers.

\section{Methodology}
\subsection{Data preparation}
We use the OptionMetrics IvyDB implied volatility, a grid of smoothed interpolated values with implied $\delta$ of the option on one axis and number of days to maturity on the other; and realized volatility calculated by OptionMetrics over $n = 28$ calendar days, using the standard deviation on the daily log return.
Following \cite{deeplearningfromivsurfaces}, we split data by year and month and drop any samples with incomplete data, producing a full dataset of 4,259,070 rows between 2012 and 2022. We augment the IV surface with the month, day, and day of the week of the observation, scaled to values in $[0,1]$, to allow the model to capture seasonal trends (Figure \ref{fig:iv_sample}). See Appendix \ref{appendix-data} for details.

\begin{figure}
    \centering
    \includegraphics[width=0.8\linewidth]{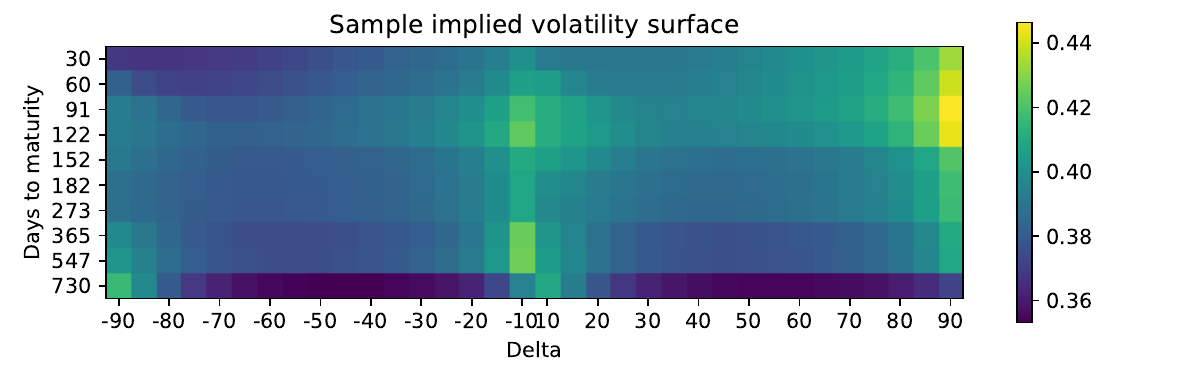}
    \caption{IV surface for NVDA stock on 2021-04-13, presented as a one-channel image instead of the traditional three-dimensional surface. Negative deltas correspond to puts.}
    \label{fig:iv_sample}
\end{figure}

\subsection{Model Architecture}

We tested both deep and wide ViT architectures, adapting the original ViT model \cite{dosovitskiy2021imageworth16x16words} for single-channel matrices of size $10 \times 36$ instead of traditional images\footnote{Model definitions and code to reproduce all results will be released with the final version of this paper.}. Because the ViT outputs a vector, we add a small four-layer MLP to produce the final real-valued prediction. We study the performance of this model on our dataset, varying the number of layers and the number of parameters per layer. Model scaling is of particular interest, as we show that small models can be trained on limited data and achieve strong performance in the task of forecasting realized volatility.
\begin{figure}[t]
    \centering
    \includegraphics[width=1.0\linewidth]{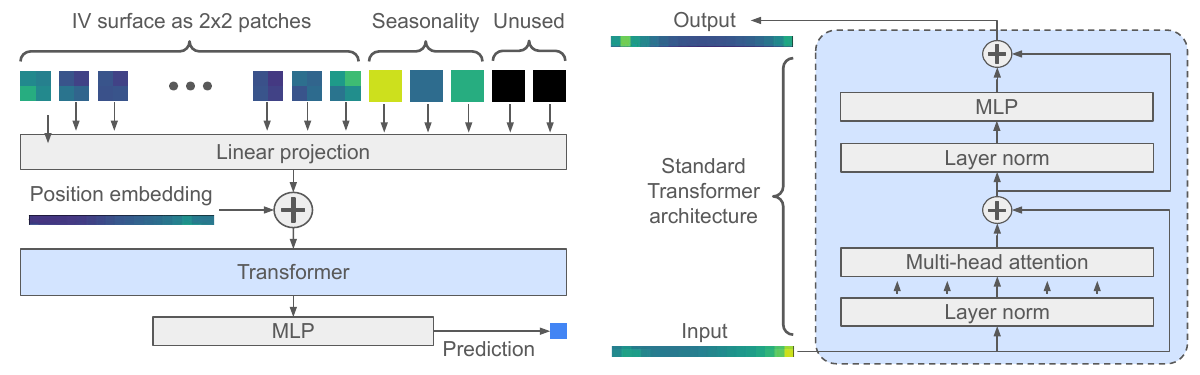}
    \caption{Vision transformer architecture (left) on our data, with more detailed schematic of the standard transformer architecture used in our model (right). In the deep Vision Transformer, the MLP layer in the Transformer model is repeated.}
    \label{fig:vision-transformer}
    \vspace{-1em}
\end{figure}
We compare our model against a baseline multilayer perceptron (MLP) on the flattened IV surface, observing that the ViT architecture outperforms the MLP. The MLP is also more difficult to train, requiring early stopping, batch normalization, and multiple training attempts with the best model selected at the end.

\begin{table}[h]
    \centering
    \begin{tabular}{c|cccc}
    \hline
    Model & \texttt{ViT\_0.005M\_wide} & \texttt{ViT\_0.12M\_deep} & \texttt{ViT\_0.17M\_wide}
    \\
    \# Params & 46466 & 122114 & 170754 \\ \hline
    Model & \texttt{ViT\_0.5M\_deep} & \texttt{ViT\_0.5M\_wide} & \texttt{ViT\_1.7M}
    \\
    \# Params & 469506 & 545282 &1732610 \\
    \hline
    \end{tabular}
    \vspace{0.5em}
    \caption{Model definitions with number of parameters; full definitions are in Appendix \ref{appendix-models}.}
    \label{tab:param-count}
    \vspace{-1.5em}
\end{table}

\subsection{Training}
 We define train and test sets such that if the test year is $y_i$, the corresponding $n$ training years are $y_{i-n}, y_{i-n+1},...,y_{i-1}$.
 We pay close attention to the choice of optimizer, learning rate schedule, and loss function to achieve more stable and efficient training. Prior work has applied early stopping, regularization, and ensembling to overcome the challenges of training on financial data \cite{empiricalassetpricing}. We apply batch normalization, a form of regularization, and Xavier initialization \cite{glorot2010understanding} in the MLP prediction component of our model. 
Our training procedure follows the process used to train text foundation models such as DeepSeek-v3 \cite{deepseekai2025deepseekv3technicalreport} and Llama \cite{grattafiori2024llama3herdmodels}, scaled down for the available data. We use a cosine annealing learning rate scheduler\footnote{\href{https://github.com/katsura-jp/pytorch-cosine-annealing-with-warmup}{https://github.com/katsura-jp/pytorch-cosine-annealing-with-warmup}, called with parameters \texttt{scheduler	scheduler\_first\_cycle\_steps = 200, scheduler\_max\_lr	= 0.01, scheduler\_min\_lr = 0.001, scheduler\_warmup\_steps = 100,	scheduler\_gamma	= 0.95}.}, introduced in \cite{loshchilov2017sgdrstochasticgradientdescent} and shown to achieve strong performance on ImageNet \cite{goyal2018accuratelargeminibatchsgd}   with large batch sizes. (We use a batch size of 2048 for all experiments.)
We use the AdamW optimizer \cite{loshchilov2019decoupledweightdecayregularization} and select the best-performing model from all training epochs. To reduce the impact of outliers, we use the Huber loss \ref{eq:huber}, a convex loss function that is quadratic for small values of $\hat y - y$ and linear for large values. We report the model's $R^2$ on unseen test data.
 \begin{equation}
     \ell_{\text{huber}}(\hat y, y) = \begin{cases}
         \frac{1}{2} \left(\hat y - y\right)^2 & |\hat y - y| \leq d
         \\
         d \cdot \left(|\hat y - y| - \frac{1}{2}d\right) & \text{otherwise}
     \end{cases}
     \label{eq:huber}
 \end{equation}

We evaluate both deep and wide ViT models of varying sizes. The deep models have four MLP layers in the Transformer module (see Figure \ref{fig:vision-transformer}), while the wide models have one MLP layer in this module.

\section{Results}
\begin{wrapfigure}{r}{0.6\textwidth}
\vspace{-2em}
    \includegraphics[width=1.0\linewidth]{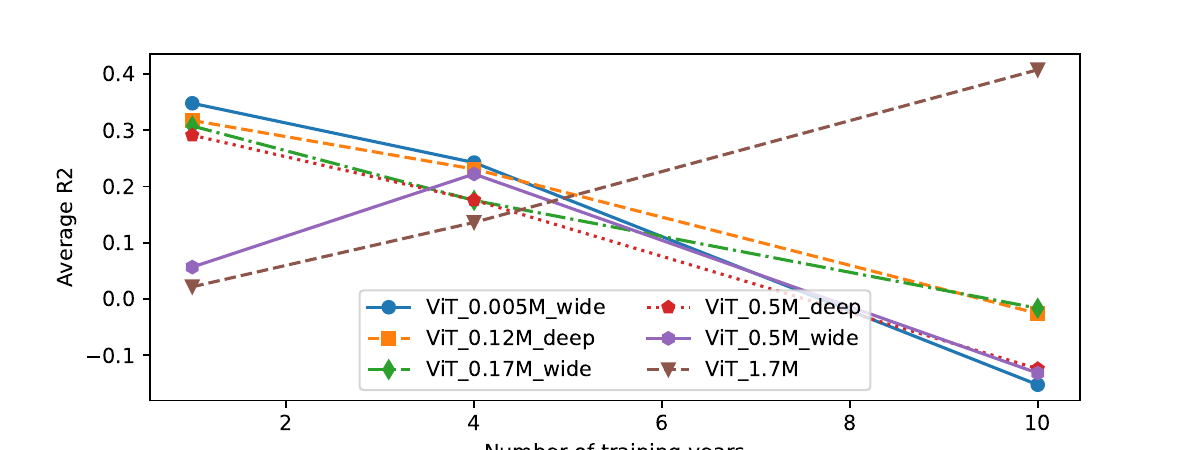}
    \caption{Effect of dataset size on model $R^2$. Where multiple train-test splits are possible, average $R^2$ is reported.}
    \label{fig:years-vs-r2}
\end{wrapfigure}
\begin{figure}

    \centering
    \includegraphics[width=0.45\linewidth]{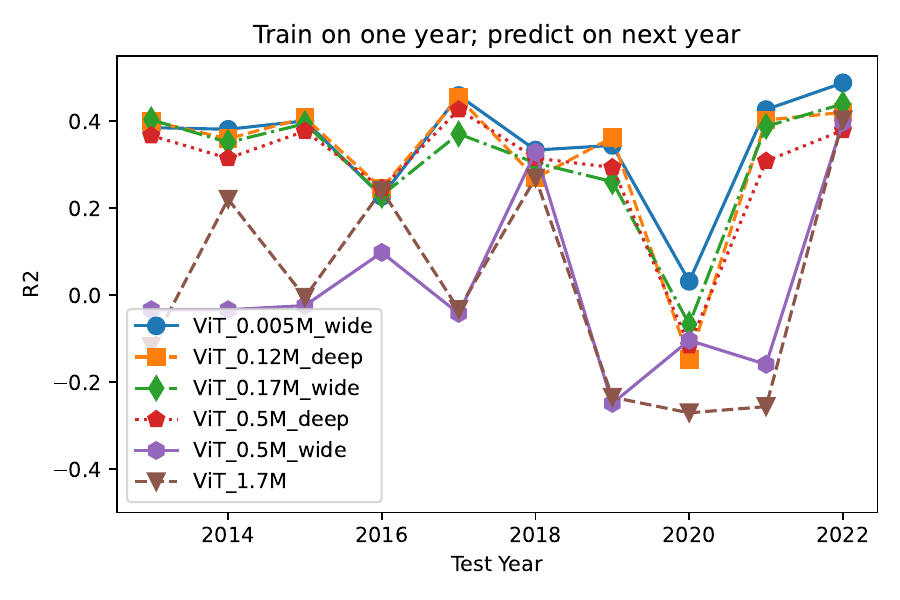}
    \includegraphics[width=0.45\linewidth]{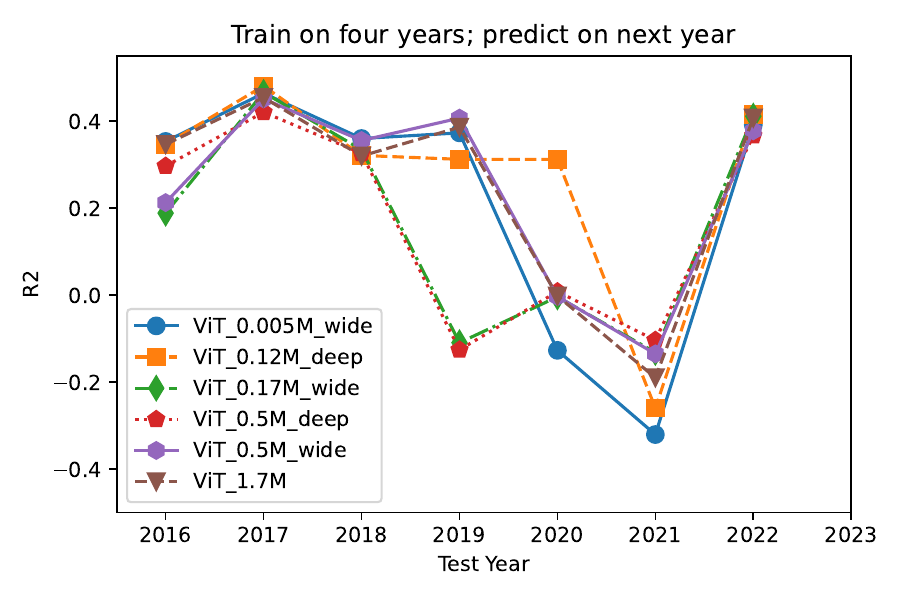}
    \caption{Training on one year (left) or four (right) and predicting the 30-day realized volatility on the next year, for data between 2012-2022. The performance drop for the test year 2020 reflects market disruption during the COVID pandemic. In practice one could iteratively retrain the models; note that the small models recover their performance on the 2021 test sample.}
    \label{fig:one-to-one}
\vspace{-1em}
\end{figure}

We study the performance of the models listed in Table \ref{tab:param-count} to predict the realized volatility of an asset over the next 30 days when trained on varying dataset sizes. Because one of the challenges for financial machine learning is the availability of data, we evaluate model performance when trained on one, four, or ten years of data, finding that smaller models (.05-.17M) can perform well when trained on smaller datasets but collapse on large ones, while the 0.5M models do not improve on the smaller models. The 1.7M model yields the best performance but requires the full ten years of training data (Figure \ref{fig:years-vs-r2}).
Additionally, all models perform poorly if the training and test data are dissimilar (Figure \ref{fig:one-to-one}). In this case, the small models provide an advantage as they could be retrained as new data becomes available.
Model performance varies across different market conditions, with all models showing reduced performance when tested on 2020 data (Figure \ref{fig:one-to-one}). 
The best model is \texttt{ViT\_1.7M} trained on 2012-2021 data, which achieves $R^2 = 0.41$ on the 2022 test set (Figure \ref{fig:years-vs-r2}). All ViT models reach their maximum $R^2$ within one or two epochs, with further training causing overfitting (Appendix \ref{appendix-results}).

Key findings: ViT models can extract nonlinear features from IV surfaces, with small models requiring as little as one year of data to train. Despite the limited availability of market data, \textbf{designing model architectures and training processes to fit the available data can enable the development of transformer models for financial forecasting tasks.}
%

\section{Ablation Testing}
We test our ViT model against two ablations: the ViT model on the IV surface with no seasonality augmentation, and an MLP-only model with roughly the same number of parameters. Full definitions of these models are provided in Appendix \ref{appendix-models}. Removing the seasonality information has a small negative effect, suggesting the model is primarily extracting nonlinear patterns from the IV surface. The MLP-only models do not perform well, with larger model sizes actually yielding worse performance.
\begin{table}[h]
    \centering
    \begin{tabular}{c|ccc}
    Model & \# Train Years & Baseline $R^2$ & No seasonality
    \\\hline
         \texttt{ViT\_0.5M\_deep} & 4 & 0.35 & 0.35
         \\
         \texttt{ViT\_0.5M\_wide} & 4 & 0.37 & 0.35
         \\
         \texttt{ViT\_1.7M} & 10 & 0.41	& 0.38
    \end{tabular}
    \vspace{0.5em}
    \caption{Effect of removing seasonality information.}
    \label{tab:seasonality}
    \vspace{-1em}
\end{table}
\begin{table}[h]
    \centering
    \begin{tabular}{c|cccccc}
    Model & ViT Params &Baseline $R^2$ & MLP Model & MLP Only $R^2$ & MLP Params
    \\\hline
    \texttt{ViT\_0.12M\_deep} & 122114& 0.27 & \texttt{MLP\_0\_12} & 0.29 & 114842
    \\
    \texttt{ViT\_0.17M\_wide} & 170754 & 0.37 & \texttt{MLP\_0\_17} & 0.29 & 174722
    \\
     \texttt{ViT\_0.5M\_deep} & 469506 & 0.35 & \texttt{MLP\_0\_5} & 0.17 & 515252
     \\
     \texttt{ViT\_0.5M\_wide} & 545282 & 	0.37 & \texttt{MLP\_0\_5} & 0.17& 515252
     \\
    \end{tabular}
    \vspace{0.5em}
    \caption{Comparison between ViT and MLP-only architectures with similar parameter counts. All models were trained on 4 years of data from 2018-2021 and tested on 2022.}  \label{tab:mlp-only}
\end{table}

\vspace{-1em}
\section{Planned and Future Work}
This is an ongoing project with many interesting directions. Our top priority is to study the potential for transfer learning on IV surfaces: the ability to fine-tune a model or retrain only the final stages of the model, such as regressor or classifier layers, to predict a different target value. ViT models trained on image datasets exhibit this property and are often fine-tuned for specific classification tasks in medical or scientific imaging \cite{transferlearning}. We are also interested in investigating whether the ViT model can learn output vectors that generalize to other prediction tasks if the MLP predictor is retrained. We tested for this capability using the task of predicting the asset's return over the next 28 days and did not observe it; however this task is more difficult than predicting the realized volatility. Following our theme of applying foundation model techniques to financial data, we could compare ensembling, applied in \cite{deeplearningfromivsurfaces}, against a Mixture-of-Experts architecture \cite{deepseekai2025deepseekv3technicalreport}. Finally, as there is clearly a link between model size, dataset size and performance, a theoretical understanding of the information content of IV surfaces could provide guidance for optimal data sampling to improve model performance.

\bibliographystyle{plainnat}
\bibliography{bibliography}
\clearpage
\appendix
\section{Data Preparation}\label{appendix-data}
\subsection{Identification of Valid Assets}
While the IV surfaces and realized volatilities are present in the OptionMetrics IvyDB dataset, we are also interested in predicting the future returns of an asset, which can be calculated using the daily returns in the CRSP (Center for Research in Security Prices) dataset. However, OptionMetrics uses the primary key \texttt{secid} and CRSP uses the primary key \texttt{cusip}. These keys do not have a one-to-one mapping because it is possible for assets to be delisted, to be added to the dataset during a calendar month, or to change primary keys (for example, due to company mergers or acquisitions). Because we batch data by month and year, we can construct a one-to-one mapping between \texttt{secid} and \texttt{cusip} using the WRDS link tables (\texttt{wrdsapps\_link\_crsp\_optionm} and the \texttt{stocknames\_\_v2} table, which provide the start and end dates during which each primary key is active. For each month of data we drop any rows where the \texttt{cusip} or \texttt{secid} is valid for only part of the month.

\subsection{Data Collection}
Using our table of valid primary keys, we downloaded raw data from OptionMetrics IvyDB, using the Volatility Surfaces and Realized Volatility tables, and the end-of-day return from the CSRP Stock dataset. The IV surface dataset contains smoothed, interpolated data on standardized calls and puts, with expirations of 10,30, 60, 91, 122, 152, 182, 273, 365, 547, and 730 calendar days, at deltas of 0.10, 0.15,0.20, 0.25, 0.30, 0.35, 0.40, 0.45, 0.50, 0.55, 0.60, 0.65, 0.70, 0.75,0.80, 0.85, 0.90 (negative deltas for puts). We fuse this data on the primary key and date, producing a total of 120 parquet files (12 months each from 2012 to 2022).

Rows are dropped if there is missing data in the IV surface or invalid values in the CRSP stock price or return values, indicating assets that did not trade on a particular day. We construct the IV surface using all available $\delta$ values and days-to-expiry for both calls and puts. We do not attempt to filter for outliers in the IV surfaces or other data.

\begin{figure}[h]
    \centering
    \includegraphics[width=0.9\linewidth]{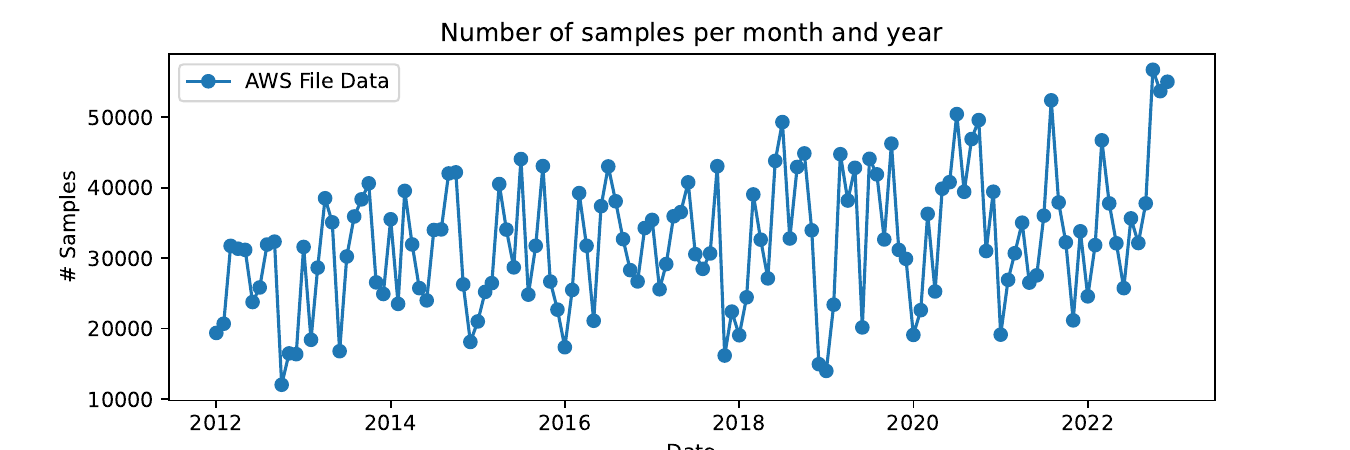}
    \caption{Number of samples per month of data.}
    \label{fig:data-rows}
\end{figure}

Our final dataset consists of 4,259,070 rows, distributed across months as shown in Figure \ref{fig:data-rows}.

\clearpage
\section{Model Definitions}
\label{appendix-models}

Table \ref{tab:model-params} lists the parameters used to initialize the various model architectures.

The Vision Transformer architecture is taken from the standard PyTorch implementation\footnote{\href{https://docs.pytorch.org/vision/main/models/vision\_transformer.html}{https://docs.pytorch.org/vision/main/models/vision\_transformer.html}}, modified to accept tensors of size $\mathbb{R}^{1 \times 10 \times 36}$ instead of square RGB images.

All models considered take input of size $10 \times 36$, operate on $2\times 2$ image patches, following the approach in \cite{deeplearningfromivsurfaces}, and produce a single real-valued prediction. Models consist of a Vision Transformer (ViT) followed by a 4-layer multilayer perceptron (MLP) to convert the ViT vector output into a single prediction.

\begin{table}[h]
    \centering
    \begin{tabular}{c|p{1cm}p{1cm}p{1cm}p{1cm}p{1cm}p{1cm}p{1cm}}
         Model & \# ViT Layers & \# Heads & \# ViT hidden dim. & ViT MLP dim. & ViT Dropout & ViT output size & MLP hidden dim. \\
         \hline
         \texttt{ViT\_0.005M\_wide} & 1 & 8 & 64 & 64 & 0.1 & 64 & 64 \\
         \texttt{ViT\_0.12M\_deep} & 4 & 8 & 64 & 64 & 0.1 & 64 & 64 \\
         \texttt{ViT\_0.17M\_wide} & 1 & 16 & 128 & 128 & 0.1 & 128 & 128 \\
         \texttt{ViT\_0.5M\_deep} & 4 & 16 & 128 & 128 & 0.1 & 128 & 128 \\
         \texttt{ViT\_0.5M\_wide} & 1 & 16 & 256 & 256 & 0.1 & 256 & 128 \\
         \texttt{ViT\_1.7M} & 4 & 16 & 256 & 256 & 0.1 & 256 & 128 \\
         \hline
    \end{tabular}
    \caption{Summary of model parameters for all model sizes.}
    \label{tab:model-params}
\end{table}

To conduct the MLP-only ablation experiment, we define the following alternate models, selected to match the parameter sizes of the ViT models.
\begin{itemize}
    \item \texttt{MLP\_0\_12}: A four-layer MLP with input size = 360 (to match the flattened IV surface) and hidden size = 180.
    \item \texttt{MLP\_0\_17}: A four-layer MLP with input size = 360 and hidden size = 240.
    \item \texttt{MLP\_0\_5}: An eight-layer MLP with input size = 360 and hidden size = 350.
\end{itemize}

\subsection{\texttt{torchinfo} summary of \texttt{MLP\_0\_5} model}
This summary was generated with an input and hidden size of 360, matching the size used in the ablation test.
\begin{verbatim}
===========================================================================
Layer (type:depth-idx)                   Output Shape              Param #
===========================================================================
DeepMLP                                  [1, 1]                    --
+-Linear: 1-1                            [1, 360]                  129,960
+-BatchNorm1d: 1-2                       [1, 360]                  720
+-ReLU: 1-3                              [1, 360]                  --
+-Linear: 1-4                            [1, 360]                  129,960
+-BatchNorm1d: 1-5                       [1, 360]                  720
+-ReLU: 1-6                              [1, 360]                  --
+-Linear: 1-7                            [1, 360]                  129,960
+-BatchNorm1d: 1-8                       [1, 360]                  720
+-ReLU: 1-9                              [1, 360]                  --
+-Linear: 1-10                           [1, 180]                  64,980
+-BatchNorm1d: 1-11                      [1, 180]                  360
+-ReLU: 1-12                             [1, 180]                  --
+-Linear: 1-13                           [1, 180]                  32,580
+-BatchNorm1d: 1-14                      [1, 180]                  360
+-ReLU: 1-15                             [1, 180]                  --
+-Linear: 1-16                           [1, 90]                   16,290
+-BatchNorm1d: 1-17                      [1, 90]                   180
+-ReLU: 1-18                             [1, 90]                   --
+-Linear: 1-19                           [1, 90]                   8,190
+-BatchNorm1d: 1-20                      [1, 90]                   180
+-ReLU: 1-21                             [1, 90]                   --
+-Linear: 1-22                           [1, 1]                    91
===========================================================================
Total params: 515,251
Trainable params: 515,251
Non-trainable params: 0
Total mult-adds (Units.MEGABYTES): 0.52
===========================================================================
\end{verbatim}

\subsection{\texttt{torchinfo} summary of all other MLP models}
This summary was generated with an input size of 256 and a hidden size of 128.
\begin{verbatim}
===========================================================================
Layer (type:depth-idx)                   Output Shape              Param #
===========================================================================
SimpleMLP                                [1, 1]                    --
+-Linear: 1-1                            [1, 128]                  32,896
+-BatchNorm1d: 1-2                       [1, 128]                  256
+-ReLU: 1-3                              [1, 128]                  --
+-Linear: 1-4                            [1, 128]                  16,512
+-BatchNorm1d: 1-5                       [1, 128]                  256
+-ReLU: 1-6                              [1, 128]                  --
+-Linear: 1-7                            [1, 64]                   8,256
+-BatchNorm1d: 1-8                       [1, 64]                   128
+-ReLU: 1-9                              [1, 64]                   --
+-Linear: 1-10                           [1, 1]                    65
===========================================================================
Total params: 58,369
Trainable params: 58,369
Non-trainable params: 0
Total mult-adds (Units.MEGABYTES): 0.06
===========================================================================
\end{verbatim}
\section{Additional Results}\label{appendix-results}

\begin{figure}
    \centering
    \includegraphics[width=1.0\linewidth]{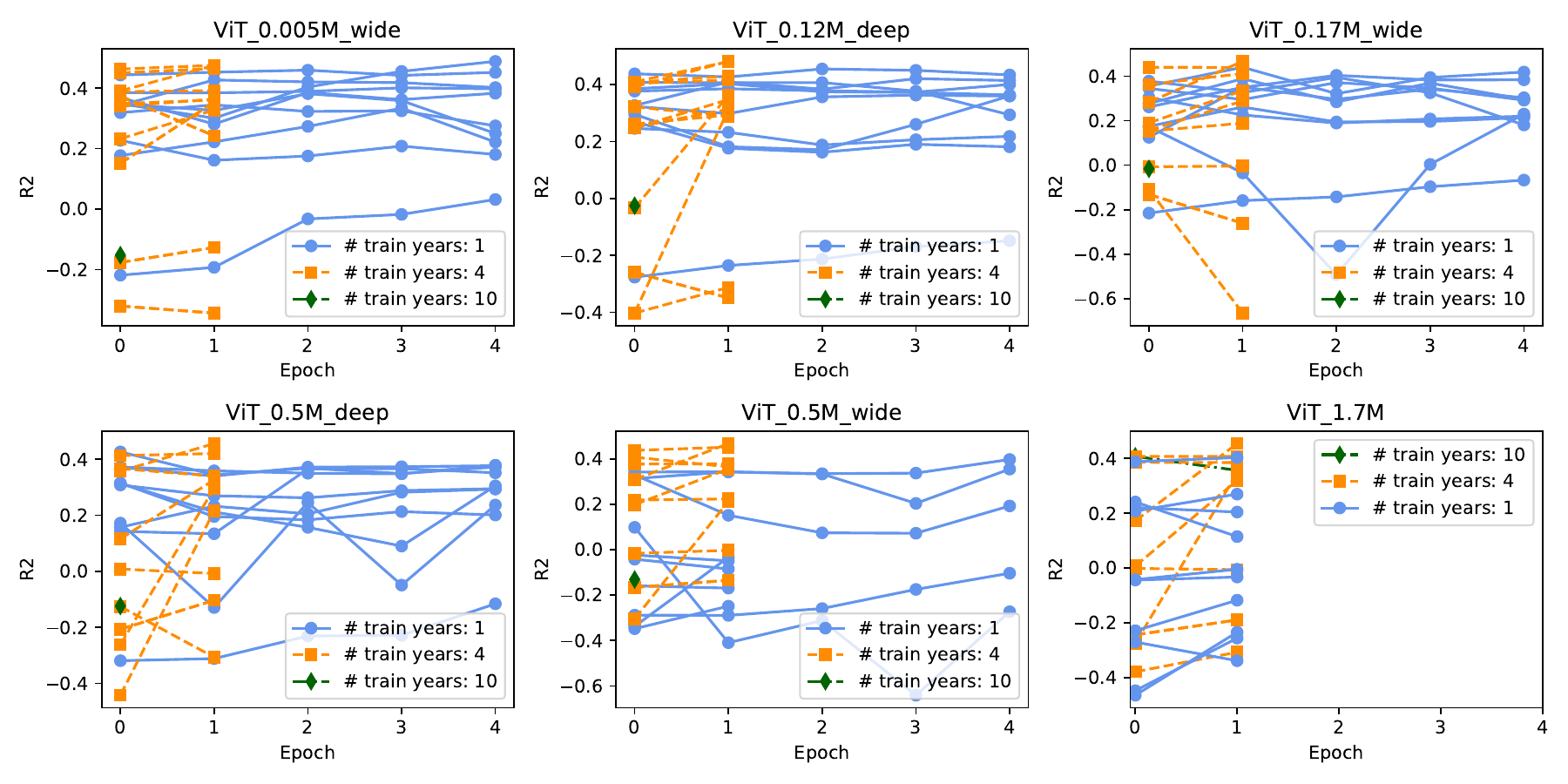}
    \caption{$R^2$ plotted over the number of training epochs, broken down by ViT model type. All of the train-on-one-year, test-on-one-year models exhibit poor performance when tested on 2020 data, corresponding to a single low or negative $R^2$ trajectory observed in each plot.}
    \label{fig:epochs-trajectories}
\end{figure}

Figure \ref{fig:epochs-trajectories} shows the training trajectories for small and large ViT models, showing how many models achieve their full performance on one or two epochs.

 We observe that while training on one year of data can produce good results, they are often inconsistent. Further, we observe the expected relationship between the number of model parameters and the data required to train the model; the smallest models perform best on small datasets, and the largest model requires the full dataset: 10 years of training data, with 1 year of test data.

\begin{figure}
    \centering
    \includegraphics[width=1.0\linewidth]{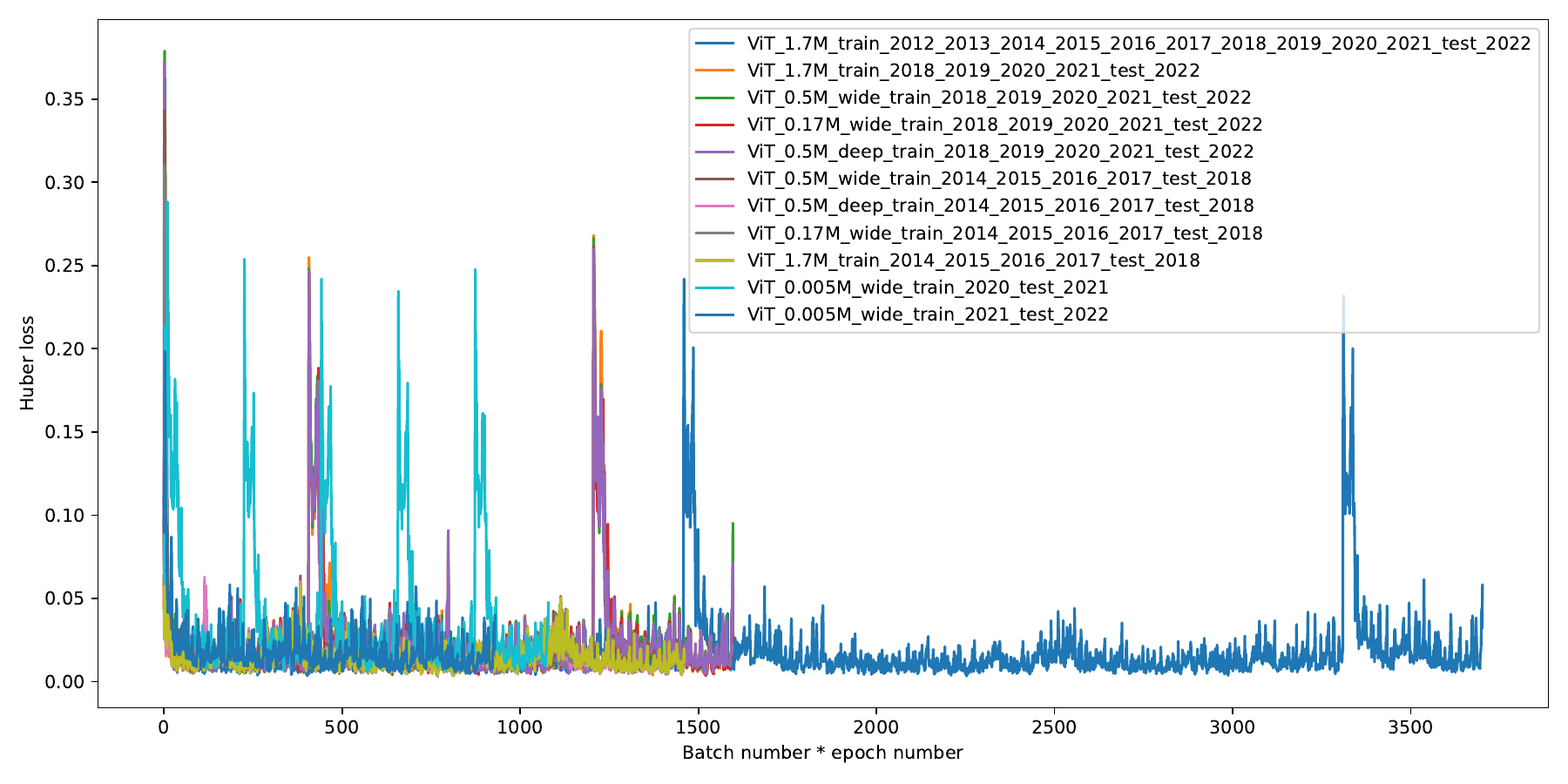}
    \\
    \includegraphics[width=1.0\linewidth]{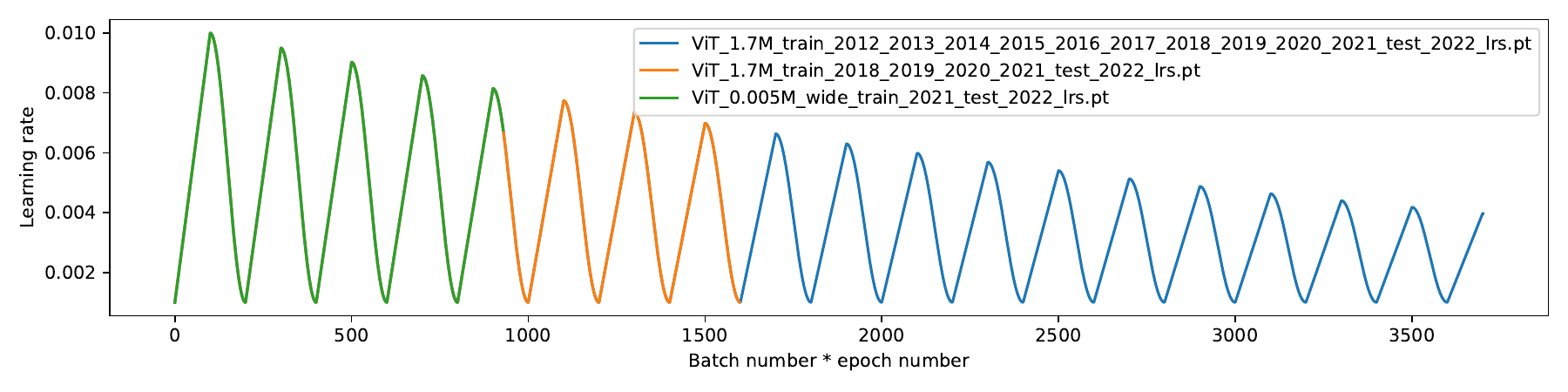}
    \caption{Huber loss and cosine learning rate for several sampled training runs.}
    \label{fig:losses}
\end{figure}

Figure \ref{fig:losses} shows the Huber loss, recorded at each batch, for some samples of 1, 4, and 10-year training runs. We observe a few loss spikes, which may be caused when the cosine learning rate increases, although in most cases the learning rate increase does not cause a spike.

\end{document}